# Conversational AI: The Science Behind the Alexa Prize

**Ashwin Ram**[1]  **Rohit Prasad**[1]  **Chandra Khatri**[1]  **Anu Venkatesh**[1]
**Raefer Gabriel**[1]  **Qing Liu**[1]  **Jeff Nunn**[1]  **Behnam Hedayatnia**[1]
**Ming Cheng**[1]  **Ashish Nagar**[1]  **Eric King**[1]  **Kate Bland**[1]
**Amanda Wartick**[1]  **Yi Pan**[1]  **Han Song**[1]  **Sk Jayadevan**[1]
**Gene Hwang**[1]  **Art Pettigrue**[1]

[1]Amazon Alexa Prize
{ashwram, roprasad, ckhatri, anuvenk}@amazon.com
{raeferg, qqliu, jeffnunn, behnam}@amazon.com
{chengmc, nashish, kinr, kateblan}@amazon.com
{warticka, yipan, hasong, skj}@amazon.com
{ehwang, pettigru}@amazon.com

## Abstract

Conversational agents are exploding in popularity. However, much work remains in the area of social conversation as well as free-form conversation over a broad range of domains and topics. To advance the state of the art in conversational AI, Amazon launched the Alexa Prize, a 2.5-million-dollar university competition where sixteen selected university teams were challenged to build conversational agents, known as "socialbots", to converse coherently and engagingly with humans on popular topics such as Sports, Politics, Entertainment, Fashion and Technology for 20 minutes. The Alexa Prize offers the academic community a unique opportunity to perform research with a live system used by millions of users. The competition provided university teams with real user conversational data at scale, along with the user-provided ratings and feedback augmented with annotations by the Alexa team. This enabled teams to effectively iterate and make improvements throughout the competition while being evaluated in real-time through live user interactions. To build their socialbots, university teams combined state-of-the-art techniques with novel strategies in the areas of Natural Language Understanding, Context Modeling, Dialog Management, Response Generation, and Knowledge Acquisition. To support the teams' efforts, the Alexa Prize team made significant scientific and engineering investments to build and improve Conversational Speech Recognition, Topic Tracking, Dialog Evaluation, Voice User Experience, and tools for traffic management and scalability. This paper outlines the advances created by the university teams as well as the Alexa Prize team to achieve the common goal of solving the problem of Conversational AI.

## 1 Introduction

Conversational AI is the study of techniques for software agents that can engage in natural conversational interactions with humans. Recently, conversational interfaces/assistants such as Amazon Alexa, Apple's Siri, Google Assistant, and others have become a focal point in both academic and industry research because of their rapid market uptake and rapidly increasing range of capabilities. The first generation of these assistants have been focused on short, task-oriented dialogs, such as playing music or asking for information, as opposed to longer free-form conversations that occur naturally in social and professional human interaction. Achieving sustained, coherent and engaging dialog is the next frontier for conversational AI.

Current state-of-the-art systems are still a long way from being able to have natural everyday conversations with humans (Levesque, 2017). There are several major challenges associated with building conversational agents: Conversational Automatic Speech Recognition (ASR) for free-form speech, Natural Language Understanding (NLU) for multi-turn dialogs, Conversational Datasets and Knowledge Ingestion, Commonsense Reasoning, Context Modeling, Dialog Planning, Response Generation, Natural Language Generation (NLG), Sentiment Detection, Inappropriate Response Filtering (e.g. profanity, inflammatory opinions, inappropriate jokes, hate speech), Personalization, Conversation Evaluation, and Conversational Experience Design.

Although conversational AI is still in its infancy, several leading university research teams are actively pushing research boundaries in this area (Weizenbaum, 1966; Sordoni et al., 2015; Vinyals et al., 2015). Access to large-scale data and real-world feedback can drive faster progress in research. To address this challenge, Amazon announced the Alexa Prize on September 26, 2016, with the goal of advancing research in Conversational AI. Through the Alexa Prize competition, participating universities were able to conduct research and test hypotheses by building socialbots were released to the general Alexa user base. Users interacted with socialbots via the "Alexa, let's chat" experience, engaged in live conversations, and left ratings and feedback for teams at the end of their conversations.

The university teams built socialbots using the Alexa Skills Kit (ASK, Kumar et al., 2017). Amazon did the ASR to convert user utterances to text and Text-To-Speech (TTS) to render text responses from socialbots in voice for the users. All intermediate steps, including NLU, Dialog Modeling, etc., were handled by socialbots, although teams were allowed to leverage the standard NLU system that is provided with ASK. Teams were also provided with live news feeds to enable their socialbots to stay current with popular topics and news events that users might want to talk about. The Alexa Prize team did conversational intent tracking, topic detection, offensive speech detection, conversational quality evaluation, traffic allocation, and scalability along with the Socialbot Invocation and Feedback Framework to enable live deployment to a large user base consisting of millions of Alexa customers.

For evaluation, we decided not to focus on the Turing Test (Turing, 1950). The goal was not to ask users to guess whether they were speaking with an AI agent or to get it to reveal itself through clever questioning. Users knew that the conversationalist at the other end was a machine; the goal, instead, was to enable users to have the best conversations that they could, conversations that were coherent, relevant, interesting, and kept users engaged as long as possible. We expect the grand challenge of 20 minute conversations will take multiple years to achieve, and the Alexa Prize was set up as a multi-year competition to enable sustained research on this problem.

Despite the difficulty of the challenge, it is extremely encouraging and satisfying to see the work the inaugural cohort of the Alexa Prize has done and what they have achieved at the end of year one of the competition. This paper describes the scientific and engineering advancements brought by university teams as well as the Alexa Prize team over the course of the year. Section 2 provides information on the conversational experience that was designed for Alexa users to interact with and provide feedback to the socialbots, as well as the developer experience for university teams participating in the competition. Sections 3 and 4 describe the software frameworks and architecture developed to support the competition. In Section 5 we describe the scientific advancements made on several problems in Conversational AI, and Section 6 provides results and findings.

## 2 Customer Experience

The Alexa Prize team focused on enhancing the customer experience (CX) for two customer groups. The first were the Alexa customers, who were the end-users for university socialbots. The second was the university teams building the socialbots. This section describes the approaches taken to anticipate, meet, and address the needs of each group.

## 2.1 CX for Alexa Users

While the Alexa Prize had clear scientific goals and objectives, Alexa users played a key role of providing feedback on the socialbots and helping us determine which socialbots were the most coherent and engaging by rating their interactions. With these users driving the direction and result of the competition, it was important for us to make sure the Alexa Prize experience was easy to interact with, that we acquired and retained a high volume of users to ensure the teams received the data points and feedback needed to improve their socialbots.

For customer experience, we didn't want users to individually enable, interact with, and keep track of 15 different socialbots. We also needed to set expectations with users that they were interacting with early-stage systems that were still in development and not necessarily as polished as a finished product would be. In addition, we wanted to be able to randomize socialbots without revealing their identity in order to receive unbiased ratings from users. To accomplish these goals, we designed and implemented the Alexa Prize skill with a natural invocation phrase that is easy to remember ("Alexa, let's chat", "Alexa, let's chat about <topic>", and common variants). The user heard a short editorial that educated them about Alexa Prize and instructed them on how to end the conversation and provide ratings and feedback. The editorial changed as needed to keep the information relevant to different phases of the competition. We kept it succinct and interesting so we would not lose the user's attention before they even had a chance to speak with a socialbot. The editorial and instructions changed as needed to keep the information relevant to the different phrases. For example, at the initial public launch on May 8, 2017, when the 15 socialbots were still in their infancy, the Alexa Prize skill started with the following editorial: "Hi! Welcome to the Alexa Prize Beta. I'll get you one of the socialbots being created by universities around the world. When you're done chatting, say stop."

After listening to the editorial, the user was handed off to one of the competing socialbots selected at random. The socialbots were certified as ASK skills and included in the Alexa Prize skill. Socialbots began the conversation with a common introduction phrase ("Hi, this is an Alexa Prize socialbot") without revealing their identity. The user could exit at any time, and was prompted to provide a verbal rating ("How do you feel about speaking with this socialbot again?") and offer additional freeform verbal feedback without knowing which socialbot they interacted with. Ratings and feedback were provided to individual teams to help them improve their socialbots.

When the Alexa Prize experience launched, we had to ensure that the socialbots received a high volume of usage data to train and improve their models despite the fact that their conversational ability was still rudimentary. There was considerable mainstream press and media following the announcement of the Alexa Prize on September 26, 2017, and every effort was made by the Alexa Prize team to maintain awareness of the challenge and the upcoming opportunity for Alexa users to become a partner in the process. Amazon promoted the competition, and Alexa Prize team worked to design exit editorials that would encourage users to provide feedback by (initially) asking them to help university teams improve their experience and (for semifinals) asking them to help evaluate socialbots and select finalists. The Alexa Prize experience was promoted through Alexa customer and developer emails, social media, blogs, and third-party publications to ensure that users remained engaged up to the Alexa Prize finals.

The carefully timed promotions resulted in high traffic during key phases of the completion. We provided load testing and scalability tools and assistance to teams to help them scale to high levels of production traffic. During the semifinals phase, which took place from July 1 to August 15, 2017, Alexa users helped determine which socialbots would advance to the final round based upon their feedback ratings. In total, users had over 40,000 hours of conversations spanning millions of interactions over the course of the competition from public launch on May 8, 2017, to finalists' code freeze on November 7, 2017.

## 2.2 CX for University Teams

Students and faculty working on the competition were key customers for Alexa Prize, and we engaged with them as developer partners. To support socialbot development, the teams were provided with unique access to Amazon Alexa resources, technologies, and personnel. The resources are outlined below.

### 2.2.1 Alexa Prize Summit

By April 2017, teams had submitted their socialbots for ASK certification after receiving signoff from their faculty advisor. Following certification, all Alexa Prize teams were invited to participate in an Alexa Prize Summit hosted at the Amazon headquarters in Seattle, WA. The summit included keynotes and deep-dive tracks on ML (machine learning) and AI, conversational experience design, marketing strategies, and access to senior Alexa leadership, scientists, engineers, designers, and marketers as well as invited speakers from Prime Air and NPR. Teams participated in 1:1 consultations on their socialbots, and attended 14 breakout sessions hosted by teams across Alexa and AWS (Amazon Web Services). Social events allowed students and faculty to further develop professional relationships with Alexa employees, our Alexa Prize team, and with each other.

### 2.2.2 Shared Resources and Tools

Data and engineering resources shared with university teams includes:

**News and current topics:**

- Washington Post Live Data API with access to news articles and reader comments, including annotated entities (people, places, organizations, and themes) from the articles.
- List of popular and current topics, generated on a nightly basis from a variety of information sources.
- CAPC (Common Alexa Prize Chats), a dataset of frequent chats (individual dialog turns) that customers have with Alexa Prize socialbots, e.g. "What's your favorite sport" or "Tell me about Game of Thrones". The chats were anonymized and aggregated across all customers, socialbots, and conversations. CAPC was limited to Alexa Prize conversations, and includes only anonymized and aggregated information across Alexa users. The dataset did not contain any customer- or socialbot-specific information, but instead represented frequent topics and interactions that users were having with Alexa Prize socialbots.

**ASR (Automatic Speech Recognition) data:**

- Access to tokenized n-best ASR hypotheses (based on our Conversational ASR model) from interactions with a socialbot, including confidence scores for each token.
- Blacklist of offensive/profane words based on our profanity detector module.

**Customer feedback data:**

- Key metrics including customer ratings of socialbot conversations, median and 90[th] percentile conversation duration, average number of dialog turns per conversation, and an anonymized leaderboard with average metrics for all socialbots.
- Transcriptions of freeform user feedback at the end of conversations with the team's socialbot.
- Hundreds of thousands of randomly selected Alexa Prize interactions annotated by trained data analysts for correctness, contextual coherence, and interestingness.
- Daily customer experience report cards during the final feedback phase, including metrics for turn-level coherence and engagement, user ratings for conversations on various topics (e.g. Movies_TV, Sports), and common user experience gaps.

**Infrastructure:**

- Free AWS services, including but not limited to:
    - GPU-based virtual machines for building models
    - SQL/NoSQL databases
    - Object-based storage with Amazon S3
- Custom ASR model tuned for conversational data.

- Custom end-pointing and extended recognition timeouts for longer free-form conversational interactions with Alexa.
- Load testing and scalability tools and architectural guidance.

**Support:**

In addition to data and infrastructure, we engaged with university teams in several ways to provide support and feedback:
- Best practices guidelines for design of engaging conversational experiences.
- An additional Amazon-only internal beta phase, to provide traffic from Amazon employees to help inform and improve socialbot performance before general availability to all Alexa users.
- Detailed report cards on initial socialbot experiences prior to public launch. These were annotated for coherence and engagement, along with the ability of socialbots to maintain anonymity of users and themselves, and handle profanity, controversial utterances and other inappropriate interactions.
- Biweekly office hours for 1:1 consultations and deep dives with a dedicated Alexa Prize Solutions Architect, Product Manager, and members of Alexa Machine Learning science teams.
- On-demand access to Alexa Prize personnel via Slack and email.

## 3  Socialbot Management Framework

### 3.1  Invocation and Feedback Framework

The "Alexa, let's chat" experience connects a user with one of the socialbots competing for the Alexa Prize. After the conversation ends, the user is asked to provide a one-to-five star rating as well as free-form voice feedback to be shared with the corresponding university team. The welcome and feedback systems are implemented as a separate Alexa skills, with handoffs between them coordinated through a framework called Links that can be used to seamlessly combine multi-stage conversational experiences. The feedback system uses a tuned ASR model to ensure high accuracy on user ratings; for semifinals, all ratings were also validated through manual transcription.

While initially we had planned to collect integer ratings, analysis of rating data showed that many users prefer to leave fractional number ratings, such as "three and a half". To improve the accuracy of rating capture, we added support for fractional numbers. Ratings and freeform feedback on each conversation were shared daily with teams to drive incremental model improvements.

### 3.2  Uptime Measurement and Availability Monitoring

Uptime measurement and availability monitoring are key to delivering a successful experience to millions of Alexa users that maintains reliability even if one or more of the socialbots is malfunctioning or offline, which was a significant concern given the complexity of many socialbot implementations and the volume of live user traffic. We developed a monitoring system that collects availability metrics on each socialbot and removes socialbots that are not responding properly to user input from the Alexa Prize experience in real time. This was achieved with a combination of passive monitoring for failure modes in user traffic, as well as active monitoring via simulated traffic to each socialbot. A notification and reminder system, with self-service reactivation, encouraged the Alexa Prize competitors to rapidly respond to any issues in their socialbot and bring them back online as soon as possible. The system also enabled us to deactivate socialbots that produced inappropriate responses. To provide time for issue resolution and testing, socialbots that were taken offline were required to wait at least 6 hours before being restored to the Alexa Prize experience.

### 3.3  Managing Customer Experience (CX)

During the early feedback phase of the competition, the quality of the 15 socialbots was highly variable. In order to ensure that users had a good experience interacting with Alexa Prize, traffic was allocated in proportion to the average user ratings over various timescales. In addition, traffic was reduced to socialbots who had difficulty in handling the large amount of traffic. We modeled and developed a weighted randomized traffic allocation system that enabled us to achieve a significantly higher average customer experience than a pure pro rata traffic allocation would provide while maintaining sufficient traffic to all socialbots to drive improvements. As socialbots improved in quality and scalability, they received more traffic. During semifinals, all socialbots received equal traffic for evaluation purposes.

We also placed proactive guardrails on topics about which socialbots could converse with Alexa users. If a user initiated a conversation about an inappropriate topic, such as sex, socialbots redirected the conversation by suggesting topics they could chat about. To mitigate risks of socialbots trained on potentially problematic data sets with profane or offensive content, such as some publicly available datasets on the Internet, we developed a system to monitor conversations for profane or offensive responses from socialbots. In such cases, the socialbot was deactivated and the notification and reminder system mentioned earlier was used to notify the team and allow them to reactivate it after they had addressed the issue. We also encouraged teams to proactively cleanse datasets and ensure that they were using sufficiently broad filters in selecting appropriate responses to users, minimizing the chance of causing offense. Overall, the frequency of such incidents remained low over the course of the competition.

While our initial content monitoring system was based on a blacklist mechanism, we also made initial strides toward building a more sensitive and contextually aware classifier to identify 1) profane content, 2) sexual content, 3) racially inflammatory content, 4) other hate speech, and 5) violent content. This system will be integrated into future Alexa Prize competitions from the outset as they are key to ensuring a positive customer experience for end users.

## 4 Architecture

This section describes architectural and design choices made by the university teams in developing their socialbots, as well as challenges pertaining to scalability. Our engineers and solutions architect worked with the teams to provide guidance on best practices, but teams were free to develop alternative approaches if they desired.

### 4.1 Summary of architectural approaches

Socialbots were implemented as Alexa skills using the Alexa Skills Kit (ASK). When a user spoke with their Alexa device, Alexa ASR was used to create a JSON-formatted payload containing the user utterance, and was sent to the socialbot's AWS Lambda endpoint for further processing (raw audio was not shared). Teams could use ASK APIs for NLU and/or develop their own. The socialbot generated a response, which was delivered back to ASK, converted to speech using Alexa TTS, and played back through the user's Alexa device. Teams took one of two approaches to generate a response to a user's utterance:

- Functions residing within AWS Lambda were used to do the dialog management and other processing to retrieve or generate a response, which was then returned to ASK for TTS
- AWS Lambda served as a proxy for the utterance and response, passing the utterance to backend processes for dialog management and retrieval or generation of a response, which was then returned to AWS Lambda and back to ASK for TTS

In both cases, teams built and trained models on GPU instances on Amazon EC2, the elastic compute infrastructure for Amazon Web Services. Teams then performed inference using the models on either Lambda or directly on an EC2 instance. For teams that used Lambda as a proxy to backend processes, the Flask micro-framework for Python (Flask, 2017) was the most common method used to receive the utterance and manage the collection and return of the response.

Many teams took a modular approach to organizing their dialog management by using ensemble of groups of one or more Amazon EC2 instances dedicated to a single task (e.g. news retrieval, facts, question-answer, weather, etc.). Responses received from these modules were then ranked by a dialog manager and returned to ASK in the form of a text response. In many cases, university teams took advantage of Amazon DynamoDB to store session and conversational state, or to provide a topic index for retrieving news or other information.

## 4.2 Scalability challenges

During the competition, multiple load tests were conducted to evaluate the scalability of socialbots, in which a sustained amount of artificial traffic was sent to each with the goal of identifying choke points within the system. We developed a load testing tool to specifically target individual socialbots at a particular rate per second for each test. Systems that exhibited the best resiliency to failure were often those that did the majority of their processing inside Lambda and not on EC2 instances. Favoring DynamoDB (a NoSQL database) over traditional SQL databases was also a common feature found in teams that were able to stay functional under increased load. One common finding during initial load testing was that while Flask made it very easy for teams to proxy requests through a web service, Flask didn't scale well in its default configuration. By using a WSGI web server like Gunicorn to serve Flask-developed services, teams were able to use multiple threads and handle a significantly higher transaction volume. Some teams employed caching clusters using Memcached or Redis to provide responses to common cached utterances without the expense of processing the utterance within their system.

# 5 Science

Many scientific advancements were made during the competition this year to advance the state of Conversational AI. Key contributions are summarized below.

## 5.1 Conversational Automatic Speech Recognition

Automatic Speech Recognition (ASR) is one of the core components of voice based assistants. Sano et al., 2017 and Hassan et al., 2015 performed an extensive analysis on identifying the causes of reformulation (users repeating if the purpose is not served in past utterance) in intelligent voice based assistants, and they found that one of the main causes of errors is ASR. Conversations, or free-form speech, do not necessarily fit a command-like structure and the space of plausible word combinations is much larger. In addition, social conversations are informal, open-ended, contain many topics and have a high out-of-vocabulary rate. Furthermore, production-grade ASR approaches must deal with a much wider array of noise and environmental conditions than the conditions in normalized research datasets often reported on in the literature in this field, particularly with in-home environments where Alexa devices are typically used. All of these challenges make Conversational ASR a challenging problem.

We developed a custom Language Model (LM) targeted specifically at open-ended conversations with socialbots and whitelisted teams to use this model for the competition. In addition to using publicly available conversational datasets such as Fisher, Switchboard, Reddit comments, (Linguistic Data Consortium) LDC news, OpenSubtitles, Yelp reviews, etc., and the Washington Post news data provided for the competition, we incrementally added hundreds of thousands of Alexa Prize utterances transcribed over the course of the competition to our dataset. Alexa Prize transcriptions were also used to tune the LM. Performance of the model improved significantly on conversational test sets over the course of the competition. Figure 1 provides the relative improvement in word error rate (WER), which reduced by nearly 33% relative to the base model over the course of the competition.

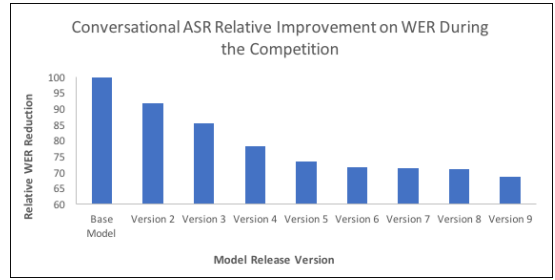

**Figure 1** Conversational ASR performance improvement: Relative reduction in WER with respect to base model.

### 5.2 Natural Language Understanding (NLU)

The following NLU components were developed during the Alexa Prize competition.

#### 5.2.1 Conversation Intent

In order to connect an Alexa user with a socialbot, we first need to identify if the user's intent is to have a conversation with Alexa. We introduced a "Conversation Intent" within the default Alexa NLU model to recognize a range of utterances such as "let's chat", "let's talk", "let's chat about <topic>", etc. using a combination of grammars and statistical models. We further expanded the experience to other natural forms of conversational initiators such as "what are we going to talk about", "can we discuss politics", "do you want to have a conversation about the Mars Mission", etc. In production, if an utterance from Alexa user is identified as "Conversation Intent", one of the Alexa Prize socialbots is invoked and the user interacts with that bot till the user says "stop".

The remaining components described below formed the core of the conversational agents and were developed by Alexa Prize teams using various approaches. The approaches are summarized here; details are available in individual papers by the teams.

#### 5.2.2 NLU Techniques for Dialogs

After the user has initiated a conversation, the socialbot requires an NLU system to identify semantic and syntactic elements from the user utterance, including user intent (such as opinion, chit-chat, knowledge, etc.), entities and topics (e.g. the entity "Mars Mission" and the topic "space travel" from the utterance: "what do you think about the Mars Mission"), user sentiment, as well as sentence structure and parse information. A certain level of understanding is needed to generate responses that align well with user intent and expectation and maximize user satisfaction.

NLU is difficult because of the complexities inherent in human language, such as anaphora, elision, ambiguity and uncertainty, which require contextual inference in order to extract the necessary information to formulate a coherent response. These problems are magnified in Conversational AI since it is an open domain problem where a conversation can be on any topic or entity and the content of the dialog can also change rapidly. Socialbot teams developed several NLU components to address these issues:

- **Named Entity Recognition (NER)**: Identifying and extracting entities (names, organizations, locations, etc.) from user utterances. Teams used various libraries such as StanfordCoreNLP (Manning et al., 2014), SpaCy (Honnibal et al., 2016), Google NLP (Google NLP API, 2017) and Alexa's ASK NLU to perform this task. NER is helpful for retrieving relevant information for response generation, as well as for tracking conversational context over multiple turns.
- **Intent Detection**: Intents represent the goal of a user for a given utterance, and the dialog system needs to detect it to act and respond appropriately to that utterance. Some of the teams built rules for intent detection or trained models in a supervised fashion by collecting the data from Amazon Mechanical Turk or by using open-source datasets such as Reddit Comments with a set of intent classes. Others utilized Alexa's ASK NLU engine for intent detection.

- **Anaphora and Co-reference Resolution**: Finding expressions that refer to the same entity in past or current utterances. Anaphora resolution is important for downstream tasks such as question answering and information extraction in multi-turn dialog systems.
- **Sentence Completion**: Some teams expanded user utterances with contextual information. For example, "Yes" can be transformed to "Yes, I love Michael Jackson" when uttered in the context of a question about the singer, or "I love Michael Jackson" can be extended to "I love Michael Jackson, singer and musician" in a conversation where this entity needs disambiguation.
- **Topic and Domain Detection**: Classifying the topic (e.g. "Seattle Seahawks") or domain (e.g. Sports) from a user utterance. Teams used various datasets to train topic detection models, including news datasets, Twitter, and Reddit comments. Some teams also collected data from Amazon Mechanical Turk to train these models.
- **Entity Linking**: Identifying information about an entity. Teams generally used publicly available knowledge bases such as Evi (Evi, 2017), FreeBase (Bollacker et al., 2008), Microsoft Concept Graph (Concept, 2017), etc. These knowledge bases can also be used to identify related entities.
- **Text Summarization**: Extracting or generating key information from documents for efficient retrieval and response generation. Some of the teams adopted this technique for efficient response generation.
- **Sentiment Detection**: Identifying user sentiment. Some teams developed sentiment detection modules to help with generating engaging responses. This can also help to better understand a user's intent and generate appropriate responses.

A robust NLU system simplifies the job for the dialog manager. For the Alexa Prize competition, the top performing teams all made extensive use of NLU capabilities.

### 5.3 Conversational Datasets, Commonsense Reasoning and Knowledge Ingestion

Currently available conversational data is limited to datasets that have been produced from online forums (e.g. Reddit), social media interactions (e.g. Twitter), and movie subtitles (e.g. Cornell Movie Dialogs (Danescu-Niculescu-Mizil and Lee, 2011), OpenSubtitles). While these datasets are useful at capturing syntactic and semantic elements of conversational interactions, many have issues with data quality, content (profanity, offensive data), tracking query-response pairs, context tracking, multiple users interacting without a specific order, and short and ephemeral conversations. To address the limitations mentioned above, we emphasized early availability of live user interactions. Through the duration of the competition, Alexa users had millions of interactions over more than 40,000 hours of conversations with socialbots. Teams used user interactions with their socialbots to evaluate and improve the quality of their socialbots iteratively.

For commonsense reasoning, several teams built modules to understand user intent. Some teams pre-processed open source and Alexa Prize datasets and extracted information about trending topics and opinions on popular topics, preparing long and short-term memory and integrating them within their dialog manager to make the responses seem as natural as possible. To complement commonsense reasoning, some of the top teams added user satisfaction modules to improve engagement along with coherence of conversations.

The teams used various knowledge bases including Amazon's Evi, Freebase, Wikidata (Wikidata, 2017), Microsoft Concept Graph, Google Knowledge Graph, and IMDb for retrieving general knowledge, facts and news, and for general question answering. Some teams also used these sources for entity linking, sentence completion and topic detection. Ideally, a socialbot should be able to ingest and update its knowledge base automatically; however, this is an unsolved problem and an active area of research. Finally, teams also ingested information from news sources such as Washington Post news and CNN to keep current with news events that users may want to chat about.

### 5.4 Dialog and Context Modeling

A key component of a conversational agent is a robust system to handle dialogs effectively. The system should accomplish two main tasks: help breakdown the complexity of the open domain problem to a manageable set of interaction modes, and be able to scale as topic diversity/breadth changes. A common strategy for dialog modeling is a hierarchical architecture with a main Dialog Manager (DM) and multiple smaller DMs corresponding to specific tasks, topics, or contexts.

Some teams, such as Sounding Board, used a hierarchical architecture and added additional modules such as an error handler to handle cases such as low-confidence ASR output or low-confidence response candidates. Other teams, such as Alquist (Pichl et al., 2017), used a structured topic-based dialog manager, where components were broken up by topics, along with intent-based dialog modules broken up by intents. Generally, teams also incorporated special-purpose modules such as a profanity/sensitivity module to filter a range of inappropriate responses and modules to address feedback and acknowledgement and to request clarity/rephrasing from users. Teams experimented with approaches to track context and dialog states, and corresponding transitions to maintain dialog flow. For example, Alquist and Slugbot (Bowden et al., 2017) modeled dialog flow as a state graph. These and other techniques helped socialbots produce coherent responses in an on-going multi-turn conversation as well as guide the direction of the conversation as needed. Few teams such as Magnus (Prabhumoye et al., 2017) even built Finite State Machines (FSMs) (Wright, 2005) for addressing specific modules such as Movies, Sports, etc. For small and static modules FSMs can be useful, however using this technique for dynamic components can be challenging as they don't scale well and context switching is harder.

The top teams focused not just on response generation but also on customer experience, and experimented with conversational strategies to increase engagement. Socialbots received an obfuscated user hashcode to enable personalization for repeat users, and some teams, such as Edina (Damonte et al., 2017), added a level of personalization to their DM to track if a certain topic or type of response is doing particularly well with a user. Few teams such e.g. Edina added engagement modules that attempted to lead the conversation through various means like news delivery, discussion on opinions, and in some cases, quizzes and voice games. Several teams (e.g. Sounding Board (Fang et al., 2017), MILABot (Serban et al., 2017), etc.) implemented a dissatisfaction and sentiment module based on user responses. Effective engagement strategies, when deployed judiciously along with response generation (see next section), context tracking and dialog flow modules, led to significant improvements in user ratings.

### 5.5 Response Generation

There are four main types of approaches for response generation in dialog systems: template/rule-based, retrieval, generative, and hybrid. A functional system can be an ensemble of these techniques, or follow a waterfall structure (e.g. rules → retrieval → generative) or use a hybrid approach with complementary modules (e.g. retrieval using generative models, generative models to create templates for a retrieval or rule-based module, etc.) Most of the teams used AIML, ELIZA (Weizenbaum, 1966), or Alicebot (Alicebot, 2017) for rule-based and templated responses. Teams also built retrieval-based modules which tried to identify an appropriate response from the dataset of dialogs available. Retrieval was performed using techniques such as N-gram matching, entity matching, or similarity based on vectors such as TF-IDF, word/sentence embeddings, skip-thought vectors, dual-encoder system, etc.

Hybrid approaches leveraging retrieval in combination with generative models are fairly new and have shown promising results in the past couple of years, usually with sequence-to-sequence approaches with some variants. Some of the Alexa Prize teams created novel techniques along these lines and demonstrated scalability and relevance for open-domain response generation deployed models in production systems. MILABot, for example, devised a Hierarchical Latent Variable Encoder-Decoder (VHRED) (Serban et al., 2016) model in addition to other Neural Network models such as Dual Encoder (Luan et al. 2016) and Skip Thought (Kiros et al. 2015) to produce hybrid retrieval-generative candidate responses. Few teams (e.g. Pixie, Adewale et al., 2017) used a two-level Long Short-Term Memory (LSTM) network (Hochreiter et al., 1997) for retrieval. EigenBot (Guss et al., 2017) and Ruby Star (Liu et al., 2017), on the other hand, used Dynamic Memory Networks (Sukhbaatar et al., 2015), Character-Level Recursive Neural Networks (RNN) (Sutskever et al., 2011) and Word-based Sequence-to-Sequence Models (Klein et al., 2017) for generating responses respectively. Alquist used a Sequence-to-Sequence model (Sutskever et al., 2014) specifically for their chit-chat module. While the above-mentioned teams deployed the generative models in production, other teams also experimented with generative and hybrid approaches offline.

**5.6 Ranking and Selection Techniques**

Open-domain social conversations do not always have a specific goal or target, and the response space can be unbounded. There may be multiple valid responses for a given utterance; identifying the response which will lead to highest customer satisfaction and help in driving the conversation forward is a challenging problem. Socialbots need mechanisms to rank possible responses and select the best response that is likely to achieve the goal of having coherent and engaging conversations. Alexa Prize teams attempted to solve this problem with either rule-based or model-based strategies.

For teams that experimented with rule-based rankers, the ranker chooses a response from the candidate responses obtained from sub-modules (e.g. topical modules, intent modules, etc.) based on some logic. For model-based strategies, teams utilized either a supervised or reinforcement learning approach, trained on user ratings (Alexa Prize data) or on pre-defined large-scale dialog datasets such as Yahoo Answers, Reddit commentaries, Washington Post Live comments, OpenSubtitles, etc. The ranker was trained to provide higher scores to correct responses (e.g. follow-up comments on Reddit are considered correct responses) while ignoring incorrect or non-coherent responses obtained by sampling. Alana (Papaioannou et al., 2017), for example, trained a ranker function on Alexa Prize ratings data and combined that with a separate ranker function that used hand-engineered features. Teams using a reinforcement learning approach developed frameworks where the agent is a ranker, the actions are the candidate responses obtained from sub-modules, and the agent is trying to maximize the tradeoff between satisfying the customer immediately versus taking into account the long-term reward of selecting a certain response. MILABot, for example, used this approach and trained a reinforcement learning ranker function on conversation ratings.

The above components form the core of socialbot dialog systems. In addition, we developed the following components to support the competition.

**5.7 Conversational Topic Tracker**

Alexa Prize conversational data is highly topical because of the nature of the social conversations. Alexa users interacted with socialbots on hundreds of thousands of topics in various domains such as Sports, Politics, Entertainment, Fashion, and Technology. This is a unique dataset collected from millions of human-conversational agent interactions. We identified the need for a conversational topic tracker for various purposes such as conversation evaluation, sentiment analysis, entity extraction, profanity detection, and response generation. To detect conversation topics in an utterance, we adopted Deep Average Networks (DAN) (Iyyer et al., 2015) and trained a topic classifier on interaction data categorized into multiple topics. We proposed a novel extension by adding topic-word attention to formulate an attention-based DAN (ADAN) (Guo et al., 2017) that allows the system to jointly capture topic keywords in an utterance and perform topic classification. We fine-tuned the model on the data collected during the course of the competition. The accuracy of the model was obtained to be 82.4% on 26 topical classes (Sports, Politics, Movies_TV, etc.). We plan to make this model available to Alexa Prize teams in the next year of the competition.

**5.8 Inappropriate and Offensive Speech Detection**

One of the most challenging aspects of delivering a positive experience to end users in socialbot interactions is obtaining high quality conversational data. The most commonly used datasets used to train dialog models are sourced from internet forums (e.g., Reddit, Twitter) or movie subtitle databases (e.g., OpenSubtitles, Cornell Movie Dialogs). These sources are all conversational in structure in that they can be transformed into utterance-response pairs. However, the tone and content of these datasets are often inappropriate for interactions between users and conversational agents, particularly when individual utterance-response pairs are taken out of context. In order to effectively use dialog models based on this or other dynamic data sources, an efficient mechanism to identify and filter different types of inappropriate and offensive speech is required.

We identified several (potentially overlapping) classes of inappropriate responses: 1) profanity, 2) sexual responses, 3) racially offensive responses, 4) hate speech 5), insulting responses, and 6) violent responses (inducements to violent acts or threatening responses). We explored keyword and pattern matching strategies, but these are subject to poor precision (with a broad list) or poor recall (with a carefully curated list). We tested a variety of Support Vector Machines and Bayesian classifiers trained on N-gram features using labeled ground truth data. The best accuracy results were in profanity (>97% at 90% recall), racially offensive responses (96% at 70% recall), and insulting responses (93% at 40% recall). More research is needed to develop effective offensive speech filters. In addition to dataset cleansing, an offensive speech classifier is also needed for online filtering of candidate socialbot responses prior to outputting them to ASK for text-to-speech conversion.

**5.9 Conversation Evaluation**

Social conversations are inherently open-ended. For example, if a user asks the question "what do you think of Barack Obama?", there are hundreds of distinct, valid, and reasonable responses. This makes training and evaluating social, non-task oriented, conversational agents extremely challenging.

The Alexa Prize competition was structured to allow users to participate in the selection of finalists. Two finalists were selected purely on the basis of user ratings averaged over all the conversations with those socialbots. In addition, one finalist was selected by Amazon based on internal evaluation of coherence and engagement of conversations by over a thousand Amazonian employees who volunteered as Alexa Prize judges, analysis of conversational metrics computed over the semifinals period, and scientific review of the team's technical papers by senior Alexa scientists.

Alexa Prize judges had conversations with the 15 socialbots using the same randomized allocation to unidentified socialbots as the general public. At the end of the conversation, they were asked to rate how coherent and engaging the conversation was. We found a correlation coefficient of 0.93 between the ranking of socialbots based on ratings provided by general Alexa users and those provided by the judges. The average rating across all socialbots was lower by 20% for the judge's pool as compared with the general public.

We also developed objective metrics (Guo et al., 2017; Venkatesh et al., 2017) for conversational quality aligned with the goals of a socialbot (the ability to converse coherently and engagingly about popular topics and current events):

  i. *Conversational User Experience (CUX):* Different users have different expectations from the socialbots, hence their experience might vary widely since open-domain dialog systems involved subjectivity. To address these issues, we used average ratings from frequent users as a metric to measure CUX. With multiple interactions frequent users have their expectations established and evaluate a socialbot in comparison to others.

 ii. *Coherence*: We annotated hundreds of thousands of randomly selected interactions for incorrect, irrelevant or inappropriate responses. Using the annotations, we calculated the response error rate (RER) for each socialbot, used to measure coherence.

iii. *Engagement*: Evaluated through performance of conversations identified as being in alignment with socialbot goals. Measured using duration, turns and ratings obtained from engagement evaluators (set of Alexa users, who were asked to evaluate socialbots based on engagement).

iv. **Domain Coverage**: Entropy analysis of conversations against the 5 socialbot domains for Alexa Prize (Sports, Politics, Entertainment, Fashion, Technology). Performance targeted on high entropy while minimizing the standard deviation of the entropy across multiple domains. High entropy ensures that the socialbot is talking about a variety of topics while a low standard deviation gives us confidence that the metric is equally well for all the domains.

v. **Topical Diversity**: Obtained using the size of topical vocabulary for each socialbot. Higher range of topics within each domain implies more topical affinity.

vi. **Conversational Depth**: We used the topical model to identify the domain for each individual utterance. Conversational depth for a socialbot was calculated as the average of the number of consecutive turns on the same topical domain. Conversational Depth evaluates socialbot's ability to have multi-turn conversations on specific topics within the five domains.

We observed that the metrics mentioned above correlate strongly with Alexa user ratings. We observed that a simple combination of the above metrics correlated strongly with Alexa user ratings (0.66), suggesting that the so-called "wisdom of crowds" (Surowiecki, 2004) is a reasonable approach to evaluation of conversational agents when conducted at large scale in a natural setting.

To automate the evaluation process, we trained a model using 60,000 conversations and ratings to predict user ratings using utterance-level and conversation-level features. Our preliminary analysis showed promising results (RMSE = 1.3) on the model trained to predict ratings from 1 to 5. We also observed strong correlation (Spearman Correlation = 0.352) between predicted ratings and corresponding user ratings for Alexa Prize conversations. These are currently the benchmark results in academia and industry. This model was not used in the competition judging process this year; however, we plan to develop it further for possible use next year.

# 6 Results

The Alexa Prize was designed as a framework to support research on Conversational AI at scale in a real-world setting. The scientific advances described above (and detailed in individual team papers) resulted in significant improvements in socialbot quality and a large amount of user engagement.

## 6.1 User Engagement

Customer engagement remained high throughout the competition. Alexa Prize ranked in the top 10 Alexa skills by usage with over 40,000 hours of conversations spanning millions of utterances. Customers chatted on a wide range of popular and current topics (26 topics) with Movies/TV, Music, Politics, Celebs, Business, and SciTech being the highest frequency (most popular) topics. Most popular topics from the post-semifinals feedback phase were Movie/TV (with average rating 3.48), SciTech (3.60), Travel/Geo (3.51), and Business (3.48). Based on user ratings, the three lowest rated topics are Arts (average rating 2.14), Shopping (2.63) and Education (3.03).

It is still early in the Alexa Prize journey towards natural human conversation, but the high level of engagement and feedback demonstrates that users are interested in chatting with socialbots and supporting their development.

## 6.2 Socialbot Quality

Over the course of the competition, socialbots showed a significant improvement in customer experience. The three finalists improved their ratings by 24.6% (from 2.77 to 3.45) over the entire duration of competition. All 15 socialbots had an average customer rating of 2.87 along with conversation duration 1:35m (median) & 5:43m (90th percentile) by the end of the semifinal phase.

Conversation duration of finalists across entire competition was 1:53 min (median) and 8:08 min (90th percentile), improving 14.0% and 56.8% respectively from competition start, with 11 turns (median) per conversation.

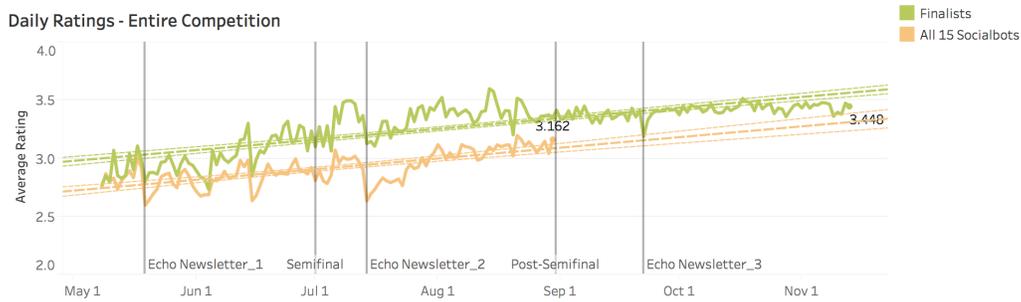

**Figure 2** Daily Ratings for socialbots during the competition.

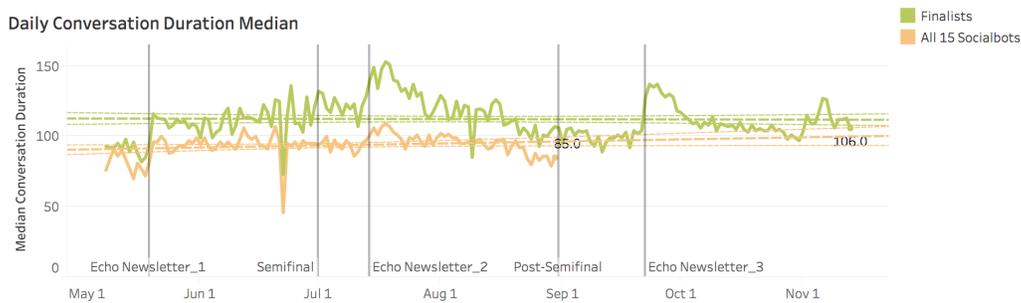

**Figure 3** Conversation Duration Median and 90th Percentile for socialbots during the competition.

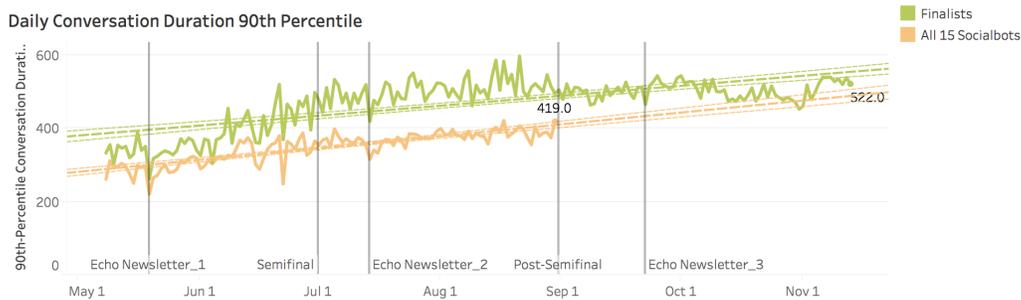

**Figure 4** Conversation Duration 90th Percentile for socialbots during the competition.

We measured Response Error Rate (RER) through manual annotation of a large fraction of user utterance-social response pairs by trained data analysts to identify incorrect, irrelevant, or inappropriate responses. RER was quite irregular during the first 30 days of the launch from May 8 to June 8, 2017, fluctuating between 8.5% and 36.7%. This was likely due to rapid experimentation by teams in response to initial user data. During semifinals from July 1 to August 15, 2017, RER was in the range of 20.8% to 28.6%. The three finalists have improved further over the post-semifinals feedback phase, and their average RER is at 11.21% (L7D) as they go into Finals.

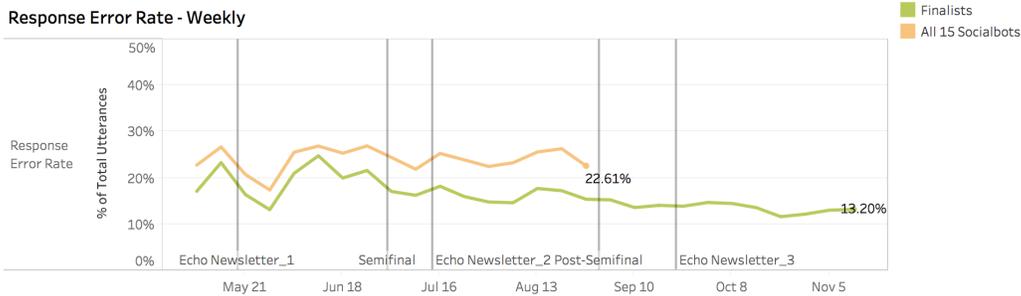

**Figure 5** Response Error Rate (RER) for individual utterance-response pairs for All Socialbots and Finalists during the competition.

## 7 Conclusion and Future Work

15 teams went live in the inaugural Alexa Prize and customer ratings improved from ~24% over the duration of the competition (May 8 to Nov 13).

Based on the work done by the teams and feedback from customers, we recommend the following key components for building an effective socialbot: 1. Dialog Manager (DM) 2. Natural Language Understanding Module (NLU) and Knowledge Module 3. Response Generation 4. Conversational User Experience (CUX) Handler 5. Ranking and Model Selection Policy Module.

We analyzed the technical and scientific detail for each team and corresponding ratings and correlated it with their performance on Socialbot quality. Following are the findings from our analysis:

- Conversational User Experience (CUX) takes least effort to build however it leads to highest gains. Building CUX is an essential component and the teams which focused most of their efforts on Science and didn't give much emphasis to CUX were not received as top performers by Alexa users.
- A robust NLU system supported by strong domain coverage leads to high coherence. Teams which invested in build a strong NLU and Knowledge component had the least Response Error Rate leading to high user ratings.
- Different conversation goals call for different response generation techniques implying that retrieval, generative and hybrid mechanisms may all be required within the same system. When the performance of a socialbot has converged, then generative and hybrid modules combined with robust ranking and selection module can lead to great gains.
- A response ranking and selection model greatly impacts socialbot quality. The teams which built a strong Model Selection Policy got significant improvement in ratings and average number of dialog turns.
- Even if a socialbot has strong response generation and ranker modules, lack of good NLU and DM, affects user ratings negatively.

All the top socialbots have strong NLU, DM and CUX modules and each one of the finalists excelled in at least one of the required components mentioned above.

We have seen significant advancement in science, and in the quality of socialbots as observed through customer ratings, but there is a lot more to be achieved. With the help of Alexa users and the science community, Alexa Prize 2018 will continue to focus on building socialbots as a vehicle to improve the state of conversational AI.

**Acknowledgments**

We would like to thank all the university students and their advisors (Alexa Prize Teams 2017) who participated in the competition. We would also like to thank Amazon leadership and Alexa

Principals for their vision and support through this entire program; Marketing, PR and Legal for helping drive the right messaging and a high volume of traffic to the Alexa Prize skill, ensuring that the participating teams received real world feedback for their research; Alexa Engineering for all the support and work on enabling the Alexa Prize skill and supporting a custom Alexa Prize ASR model while always maintaining operational excellence; and Alexa Machine Learning for continued support with NLU and data services, which allowed us to capture user requests to initiate conversations and also provide high quality annotated feedback to the teams. We also want to thank ASK Leadership and the countless teams in ASK who helped us with the custom APIs for Alexa Prize teams, enabling skill beta testing for the Alexa Prize skills before it went GA, skill management, QA, certification, marketing, operations and solutions support. We'd also like to thank the Alexa Experiences org for exemplifying customer obsession by providing us with critical input to share with the teams on building the best customer experiences and driving us to track our progress against customer feedback.

And last but not least, thank you to the Alexa customers who engaged in over 40,000 hours of conversations spanning millions of interactions with the Alexa Prize socialbots and provided the feedback that helped them improve over the course of the year.

**References**

Hector J Levesque. 2017. Common Sense, the Turing Test, and the Quest for Real AI: Reflections on Natural and Artificial Intelligence. MIT Press.

Joseph Weizenbaum. Eliza – a computer program for the study of natural language communication between man and machine. Commun. ACM , 9(1):36–45, January 1966.

Alessandro Sordoni, Michel Galley, Michael Auli, Chris Brockett, Yangfeng Ji, Margaret Mitchell, Jian-Yun Nie, Jianfeng Gao, and Bill Dolan. A neural network approach to context sensitive generation of conversational responses. In Proc. NAACL , pages 196–205, 2015.

Oriol Vinyals and Quoc V. Le. A neural conversational model. In Proc. ICML, 2015.

AlexaPrizeTeams. 2017. The Alexa Prize Teams the socialbot competition teams. https://developer.amazon.com/alexaprize. [Accessed: 2017-10-28].

Anjishnu Kumar, Arpit Gupta, Julian Chan, Sam Tucker, Bjorn Hoffmeister, Markus Dreyer. Just ASK: Building an Architecture for Extensible Self-Service Spoken Language Understanding. arXiv preprint arXiv:1711.00549

Alan M Turing. 1950. Computing machinery and intelligence. Mind 59(236):433–460.

Shumpei Sano, Nobuhiro Kaji, Manabu Sassano. Predicting Causes of Reformulation in Intelligent Assistants. arXiv preprint arXiv:1707.03968

Ahmed Hassan, Ranjitha Gurunath Kulkarni, Umut Ozertem, and Rosie Jones. 2015. Characterizing and predicting voice query reformulation. In Proceedings of the 24th ACM International on Conference on Information and Knowledge Management. ACM, pages 543–552. https://doi.org/10.1145/2806416.2806491.

Christopher D. Manning, Mihai Surdeanu, John Bauer, Jenny Finkel, Steven J. Bethard, and David McClosky. The Stanford CoreNLP natural language processing toolkit. In Proc. ACL: System Demonstrations , pages 55–60, 2014.

Honnibal, M. (2016). SpaCy (Version 1.3.0). Retrieved from: https://spacy.io/

Google Natural Language API https://cloud.google.com/natural-language. [Accessed: 2017-10-28].

EVI Knowledge API. https://www.evi.com/about/ [Accessed: 2017-10-28].

Kurt Bollacker, Colin Evans, Praveen Paritosh, Tim Sturge, and Jamie Taylor. 2008. Freebase: a collaboratively created graph database for structuring human knowledge. In Proceedings of the 2008 ACM SIGMOD International Conference on Management of Data, pages 1247–1250.

Microsoft Concept Graph API. https://concept.research.microsoft.com/Home/Introduction [Accessed: 2017-10-28].

WikiData API https://www.wikidata.org/wiki/Wikidata:Main_Page [Accessed: 2017-10-28].

Wright, David R. (2005). "Finite State Machines" (PDF). CSC215 Class Notes. David R. Wright website, N. Carolina State Univ. Retrieved July 14, 2012.

Alicebot http://www.alicebot.org/aiml.html [Accessed: 2017-10-28].

Iulian Vlad Serban, Alessandro Sordoni, Ryan Lowe, Laurent Charlin, Joelle Pineau, Aaron Courville, Yoshua Bengio. A Hierarchical Latent Variable Encoder-Decoder Model for Generating Dialogues. arXiv preprint arXiv:1605.06069

Yi Luan, Yangfeng Ji, and Mari Ostendorf. Lstm based conversation models. arXiv preprint arXiv:1603.09457 , 2016.

R. Kiros, Y. Zhu, R. R. Salakhutdinov, R. Zemel, R. Urtasun, A. Torralba, and S. Fidler. Skip-thought vectors. In NIPS, 2015.

Sepp Hochreiter and Jürgen Schmidhuber. Long short-term memory. Neural computation , 9(8):1735–1780, 1997.

Sainbayar Sukhbaatar, Arthur Szlam, Jason Weston, and Rob Fergus. Weakly supervised memory networks. CoRR , abs/1503.08895, 2015. URL http://arxiv.org/abs/1503.08895 .

M. Iyyer, V. Manjunatha, J. Boyd-Graber, and H. Daumé III. Deep unordered composition rivals syntactic methods for text classification. In Proceedings of ACL , 2015.

Fenfei Guo, Angeliki Metallinou, Chandra Khatri, Anirudh Raju, Anu Venkatesh, Ashwin Ram. Topic-based Evaluation for Conversational Bots. In NIPS, 2017.

Anu Venkatesh, Chandra Khatri, Ashwin Ram, Fenfei Guo, Raefer Gabriel, Ashish Nagar, Rohit Prasad, Ming Cheng, Behnam Hedayatnia, Angeliki Metallinou, Rahul Goel, Shaohua Yang, Anirudh Raju. On Evaluating and Comparing Conversational Agents. In NIPS, 2017.

James Surowiecki. The wisdom of crowds: Why the many are smarter than the few and how collective wisdom shapes business. J Surowiecki - Economies, Societies and Nations, 2004 - Little, Brown London

Flask Python Tool. http://flask.pocoo.org/ [Accessed: 2017-10-28].

Cristian Danescu-Niculescu-Mizil and Lillian Lee. Chameleons in imagined conversations: A new approach to understanding coordination of linguistic style in dialogs. In Proceedings of the Workshop on Cognitive Modeling and Computational Linguistics, ACL 2011, 2011.

Ilya Sutskever, JamesMartens, and Geoffrey E Hinton. Generating text with recurrent neural networks. In Proceedings of the 28th International Conference on Machine Learning (ICML-11), pages 1017–1024, 2011.

Guillaume Klein, Yoon Kim, Yuntian Deng, Jean Senellart, and Alexander M Rush. Opennmt: Open-source toolkit for neural machine translation. arXiv preprint arXiv:1701.02810, 2017.

Sutskever, I.; Vinyals, O.; Le, Q. V. Sequence to sequence learning with neural networks. In Advances in neural information processing systems, 2014, pp. 3104–3112.

Jan Pichl, Petr Marek, Jakub Konrád, Martin Matulík, Hoang Long Nguyen, Jan Šedivý. Alquist: The Alexa Prize Socialbot. Alexa Prize Proceedings, 2017.

Kevin K. Bowden, JiaqiWu, Shereen Oraby, Amita Misra, and Marilyn Walker. Slugbot: An Application of a Novel and Scalable Open Domain Socialbot Framework. Alexa Prize Proceedings, 2017.

Shrimai Prabhumoye, Fadi Botros, Khyathi Chandu, Samridhi Choudhary, Esha Keni, Chaitanya Malaviya, Thomas Manzini, Rama Pasumarthi, Shivani Poddar, Abhilasha Ravichander, Zhou Yu, Alan Black. Building CMU Magnus from User Feedback. Alexa Prize Proceedings, 2017.

Ben Krause Marco Damonte, Mihai Dobre, Daniel Duma, Joachim Fainberg, Federico Fancellu, Emmanuel Kahembwe, Jianpeng Cheng, Bonnie Webberz. Edina: Building an Open Domain Socialbot with Self-dialogues. Alexa Prize Proceedings, 2017.

Iulian V. Serban, Chinnadhurai Sankar, Saizheng Zhang, Zhouhan Lin, Sandeep Subramanian, Taesup Kim, Sarath Chandar, Nan Rosemary Ke, Sai Mudumba, Alexandre de Brebisson, Jose M. R. Sotelo, Dendi Suhubdy, Vincent Michalski, Alex Nguyen and Yoshua Bengio. The Octopus Approach to the Alexa Competition: A Deep Ensemble-based Socialbot. Alexa Prize Proceedings, 2017.

Hao Fang, Hao Cheng, Elizabeth Clark, Ariel Holtzman, Maarten Sap, Mari Ostendorf, Yejin Choi, and Noah A. Smith. Sounding Board – University of Washington's Alexa Prize Submission. Alexa Prize Proceedings, 2017.

Oluwatosin Adewale, Alex Beatson, Davit Buniatyan, Jason Ge, Mikhail Khodak, Holden Lee, Niranjani Prasad, Nikunj Saunshi, Ari Seff, Karan Singh, Daniel Suo, Cyril Zhang, Sanjeev Arora. Pixie: A Social Chatbot. Alexa Prize Proceedings, 2017.

William H. Guss, James Bartlett, Phillip Kuznetsov, Piyush Patil. Eigen: A Step Towards Conversational AI. Alexa Prize Proceedings, 2017.

Huiting Liu, Tao Lin, Hanfei Sun, Weijian Lin, Chih-Wei Chang, Teng Zhong, Alexander Rudnicky. RubyStar: A Non-Task-Oriented Mixture Model Dialog System. Alexa Prize Proceedings, 2017.

Ioannis Papaioannou, Amanda Cercas Curry, Jose L. Part, Igor Shalyminov, Xinnuo Xu, Yanchao Yu, Ondrej Dušek, Verena Rieser, Oliver Lemon. Alana: Social Dialogue using an Ensemble Model and a Ranker trained on User Feedback. Alexa Prize Proceedings, 2017.